\documentclass[runningheads]{llncs}
\usepackage{float}
\usepackage{subfig}
\usepackage{graphicx}
\usepackage{amsmath}

\begin{document}
	\title{Precise localization within the GI tract by combining classification of CNNs and time-series analysis of HMMs.\thanks{This work has been partly funded by the German Federal
Ministry of Education and Research (BMBF) in the project MEDGE (16ME0530)}}

	\titlerunning{GI tract localization with CNNs and time-series analysis}
	\author{Julia Werner\inst{1} \and
		Christoph Gerum\inst{1}\and
		Moritz Reiber\inst{1}\and
        Jörg Nick\inst{2}\and
		Oliver Bringmann\inst{1}}
	\authorrunning{Werner et al.}

	\institute{Department of Computer Science, University of Tübingen, Germany.
 \and Department of Mathematics, ETH Zürich, Switzerland.}
	\maketitle              % typeset the header of the contribution
	\begin{abstract}
		This paper presents a method to efficiently classify the gastroenterologic section of images derived from Video Capsule Endoscopy (VCE) studies by exploring the combination of a Convolutional Neural Network (CNN) for classification with the time-series analysis properties of a Hidden Markov Model (HMM). It is demonstrated that successive time-series analysis identifies and corrects errors in the CNN output. Our approach achieves an accuracy of $98.04\%$ on the Rhode Island (RI) Gastroenterology dataset. This allows for precise localization within the gastrointestinal (GI) tract while requiring only approximately $1$M parameters and  thus, provides a method suitable for low power devices.
		%The abstract should briefly summarize the contents of the paper in
		%15--250 words.
		
	\end{abstract}
	\section{Introduction}
	The capsule endoscopy is a medical procedure that has been used for investigating the midsection of the GI tract since early 2000~\cite{iddan2000wireless,costamagna2002prospective}. This minimally invasive method allows to visualize the small intestine, which is in most part not accessible through standard techniques using flexible endoscopes~\cite{smedsrud2021kvasir}. The procedure starts by swallowing a pill-sized capsule. While it moves through the GI tract by peristalsis, it sends captured images from an integrated camera with either an adaptive or a defined frame rate to an electronic device. The overall aim of this procedure is to detect diseases affecting the small intestine such as tumors and its preliminary stages, angiectasias as well as chronic diseases~\cite{smedsrud2021kvasir,mclaughlin2013primary,thomson2001small}.
	Since the esophagus, stomach and colon can be more easily assessed by standard techniques, the small intestine section is of main interest in VCE studies. 
	
	All images of the small intestine should be transmitted for further evaluation by medical experts who are qualified to check for anomalies. The frame rate of the most prominent capsules ranges from $1$ to $30$ frames per second  with a varying resolution between $256\times256$ and $512\times512$ depending on the platform~\cite{smedsrud2021kvasir}. For example, the PillCam® SB3 by Medtronic lasts up to $12$ hours with an adaptive frame rate of $2$ to $6$ frames per second \cite{monteiro2016pillcam}. This should ensure passing through the whole GI tract before the energy of the capsule's battery is depleted. However, a capsule can also require more than one day to pass through the whole GI tract leading to an incomplete record of images due to depletion of the capsule's battery after maximal $12$ hours. In this procedure, the energy is the bottleneck and small changes of the architecture can increase the overall energy requirement leading to a shorter battery lifetime with the risk of running out of energy without covering the small intestine. However, modifications such as capturing images with a higher resolution might improve the recognition ability of clinicians and thus, it is desirable to increase the limited resolution or add more functions (e.g. zooming in or out, anomaly detection on-site) helping to successfully scan the GI tract for anomalies at the cost of increasing energy demands. The images taken before the small intestine are not of interest but demand their share of energy for capturing and transmitting the images.

	This paper presents a method for very accurately determining the location of the capsule by on-site evaluation using a combination of neural network classification and time-series analysis by a HMM. This neglects the necessity to consume electric energy for transmitting images of no interest. If this approach is integrated into the capsule it can perform precise self-localization and the transition from the stomach to the small intestine is verified with high confidence. From this moment onwards, all frames should be send out for further evaluation. A major part of the energy can be saved since the data transmission only starts after the capsule enters the small intestine and therefore can be used for other valuable tasks. For example, the frame rate or resolution could be increased while in the small intestine or additionally, a more complex network for detecting anomalies on-site could be employed.

	\subsection{Related work}
    In the field of gastroenterology, there have been different approaches to perform localization of a capsule within the GI tract \cite{mateen2017localization} including but not limited to magnetic tracking~\cite{pham2014real,yim20133}, video-based \cite{zhou2014measurement,marya2014computerized}  and electromagnetic wave techniques \cite{zhang2010design,goh2014doa}. However, Charoen et al~\cite{charoen2022rhode} were the first to publish a dataset with millions of images classified into the different sections of the GI tract. They achieved an accuracy of $97.1\%$ with an  Inception ResNet V2~\cite{szegedy2017inception} architecture on the RI dataset and therefore successfully demonstrated precise localization without aiming for an efficient realization on hardware. To the best of our knowledge, there is no superior result than the baseline with this dataset. However, a large network with $56$M parameters as the Inception ResNet V2 is not suitable for low-power embedded systems since the accompanied high energy demand results in a short battery lifetime. Thus, we present a new approach for this problem setting using the same dataset and the same split resulting in a higher accuracy while requiring a much smaller network and less parameters.

	\section{Methodology}
	\subsection{Inference}
    To improve the diagnosis within the GI tract, a tool for accurate self-localization of the capsule is presented. Since the energy limitation of such a small device needs to be considered, it is crucial to limit the size of the typical large deep neural network. The presented approach achieves this by improving the classification results of a relatively small CNN with subsequent time-series analysis. 
    
     CNNs have been successfully used for many different domains such as computer vision, speech and pattern recognition tasks~\cite{abdel2014convolutional,he2016deep,huang2017densely} and thus, were employed for the classification task in this work. MobileNets~\cite{howard2017mobilenets} can be categorized as leight weight CNNs, which have been used for recognition tasks while being efficiently employed on mobile devices. Since a low model complexity is essential for the capsule application, the MobileNetV3-Small~\cite{howard2019searching} was utilized and is in the following interchangeably referred to as CNN. For subsequent time-series analysis, a HMM was chosen, since the statistical model is well established in the context of time series data \cite{rabiner1986introduction}. Due to the natural structure of the GI tract in humans, the order of states within a VCE is known. The capsule traverses the esophagus, stomach, small intestine and colon sequentially in every study. This inherent structure of visited locations can be directly encoded into the transition probabilities of the HMM. 
     Finally, the predictions from the CNN are interpreted as the emissions of a HMM and the Viterbi algorithm \cite{forney1973viterbi} is used to compute the most likely sequence of exact locations, given the classifications of the CNN.
    	
	\begin{figure}[h!]
		\includegraphics[width=0.7\textwidth]{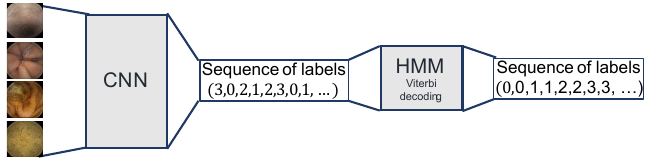}
		\centering
		\caption{Illustration of the presented approach (GI tract images from \cite{charoen2022rhode}).}
		\label{fig:illustrations}
	\end{figure}

    Hence, the presented method for localizing the four different gastroenterology sections consists of two phases as depicted in Figure~\ref{fig:illustrations}. For each patient, the CNN classifies chronologically received input data from the RI gastroenterology VCE dataset~\cite{charoen2022rhode} into the four given classes. The respective output labels/predictions of the CNN are fed into a HMM, which uses the Viterbi algorithm for determining the most likely sequence of states given the observations from the CNN. With four hidden states and often much more than $10 000$ observations per patient, the size of the matrix storing the likelihood values for each hidden state and each observation has usually more than $40 000$ entries. However, the less storage is required, the more useful this method becomes for low power devices. Furthermore, as the decoding is performed backwards a larger matrix leads to an increasing delay in classification. Thus, to limit the size of the matrix, a sliding window of size $n$ was used to build the matrix with a predefined shape. After succeeding the $n^{\text{th}}$ observation, for each new addition to the matrix the first column is removed, ensuring the predefined shape. The designated route the capsule moves along is known by the given anatomy of the GI tract. Therefore, specific assumptions can be made confidentially, e.g. the capsule cannot simply skip an organ, nor does the capsule typically move backwards to an already passed organ. To exploit this advantage of prior knowledge, the Viterbi decoding is used to detect the transitions for each organ by limiting the possible transition to the subsequent organ until a transition is detected.
	
	\subsection{HMM and Viterbi decoding} 
    HMMs are popular tools in the context of time-dependent data, e.g. in pattern recognition tasks~\cite{rabiner1989tutorial,kenny1990linear,trentin2003robust}, characterized by low complexity compared to other models. The probabilistic modeling technique of a HMM assumes an underlying Markov chain, which describes the dynamic of the hidden states $\{S_1,\dots,S_n\}$. In the present setting, the exact location of the capsule is interpreted as the hidden state, which gives $\{S_1=\text{Esophagus}, S_2=\text{Stomach}, S_3=\text{Small intestine}, S_4=\text{Colon}\}$. At any time point $t$, an emission $X_t\in\{K_1,\dots, K_m \}$, corresponding to one of the locations as classified by the neural network, is observed and assumed to be sampled from the emission probabilities, which only depend on the hidden state $S_t$. The model is thus completely determined by the transition probabilities $a_{ij}$ of the Markov chain $S_t$ (as well as its initial distribution $\pi_i=P(X_1=S_i)$ for $i\le 4$) and the emission probabilities $b_j(k)$, which are given by the probabilities
    \begin{align*}
     a_{ij}=P(X_{t+1} =S_j | X_t = S_i), \,
    \quad b_{j}(k) = P(O_t = K_k | X_t = S_j ),
    \end{align*}
    for $i,j\le 4$ and $k\le 4$, where $O_t$ denotes the observation at time $t$ \cite{manning1999foundations,eddy1996hidden}. The objective of the model is then to infer the most likely hidden states, given observations $O=(O_1,\dots,O_t)$, which is effectively realized by the Viterbi algorithm~\cite{forney1973viterbi}.  
    The Viterbi algorithm then efficiently computes the most likely sequence of gastroenterologic states $X=(X_1,\dots, X_t)$, given the evaluations of the neural network $(O_1,\dots,O_t)$, namely
    \begin{align*}
        \arg \max\limits_{X_1,\dots,X_{t}} P(X_1,\dots, X_{t}, O_1, ... O_{t}).
    \end{align*}

    To determine good approximations for the transition and emission probabilities in this problem setting, a grid search was performed. For the diagonal and superdiagonal entries of the probability matrix, a defined number of values was tested as different combinations and for each variation the average accuracy computed for all patients (all other values were set to zero). The final probabilities were then chosen based on the obtained accuracies from the grid search and implemented for all following experiments. An additional metric is employed to evaluate the time lag between the first detection of an image originating from the small intestine and the actual passing of the capsule at this position. This delay arises due to the required backtrace of the matrix storing the log-likelihoods during the Viterbi decoding before the final classification can be performed.
	
	\section{Results and Discussion}
\begin{figure}[h!]
        \centering
		\subfloat[Average delays and accuracies.]{\includegraphics[width=0.49\textwidth]{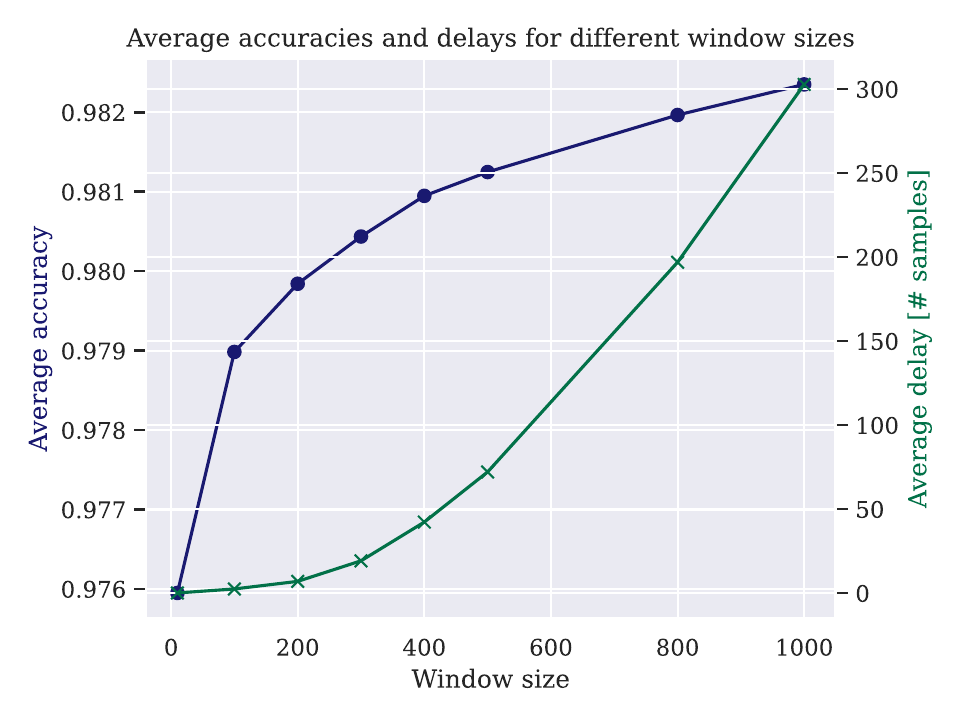}}
		\hfill
		\subfloat[Distribution of delays.]{\includegraphics[width=0.49\textwidth]{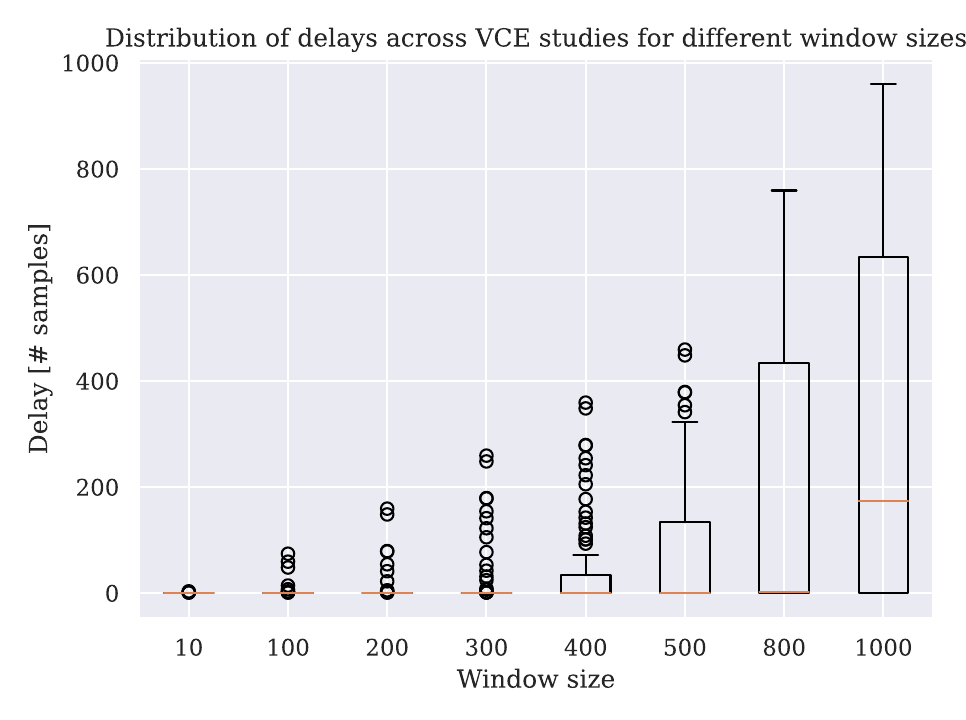}\label{fig:boxplot}}
		\caption{Delays and accuracies for different window sizes sliding over the log-likelihood matrix of the Viterbi decoding.}
        \label{fig:window}
	\end{figure}

    To determine the window size used during Viterbi decoding for subsequent experiments, the average accuracy as well as the average delays of the Viterbi decoding after classification with the CNN+HMM combination are plotted over different window sizes (see Figure~\ref{fig:window}). A larger window size results in a higher accuracy and a larger delay, while a window size reduction leads to an accuracy loss but also a decrease of the delay. A reasonable tradeoff seems to be given by a window size of $300$ samples, which was chosen for further experiments.

    Table~\ref{tab2} displays the results of the CNN+HMM combination in comparison to only using the CNN on different input image sizes (averaged describes the average over the values per patient). The $n \times n$ center of the input image was cropped and used as an input to explore the reduced complexity in terms of accuracy. This demonstrates that combining the CNN with time-series analysis can compensate a proportion of false classifications of the CNN and enhances its overall classification abilities notably. However, for all subsequent experiments an input image size of $320 \times 320$ was used for better comparability with the results of the original authors.

    \begin{table}[h!]
			\caption{Results of the CNN+HMM approach with different input image sizes.}\label{tab2}\centering
			\begin{tabular}{c|c|c|c|c|c|c}
					\hline
                    Input size & \multicolumn{2}{c|}{$64 \times 64$} & \multicolumn{2}{c|}{$120 \times 120$} & \multicolumn{2}{c}{$320 \times 320$}\\\hline
					Metric &  CNN & CNN+HMM & CNN & CNN+HMM & CNN & CNN+HMM\\
					\hline
					Accuracy [$\%$] & $90.60$& $96.16$& $93.94$& $97.37$ &  $96.95$ &$98.04$ \\
					Averaged MAE & $0.1178$ & $0.0560$  &$0.07895$& $0.0406$ & $0.0463$ & $0.0350$\\
					Averaged R2-Score & $0.1764$ &$ 0.6889$ & $0.4454$ & $0.7819$ & $0.7216$ & $0.8077$\\
					Average Delay  & --  & $19.11$& -- & $17.87$ & -- & $19.19$\\
					\hline
				\end{tabular}
		\end{table}

	\begin{figure}[h!]
		\centering
		\subfloat[CNN output.]{\includegraphics[width=0.47\textwidth]{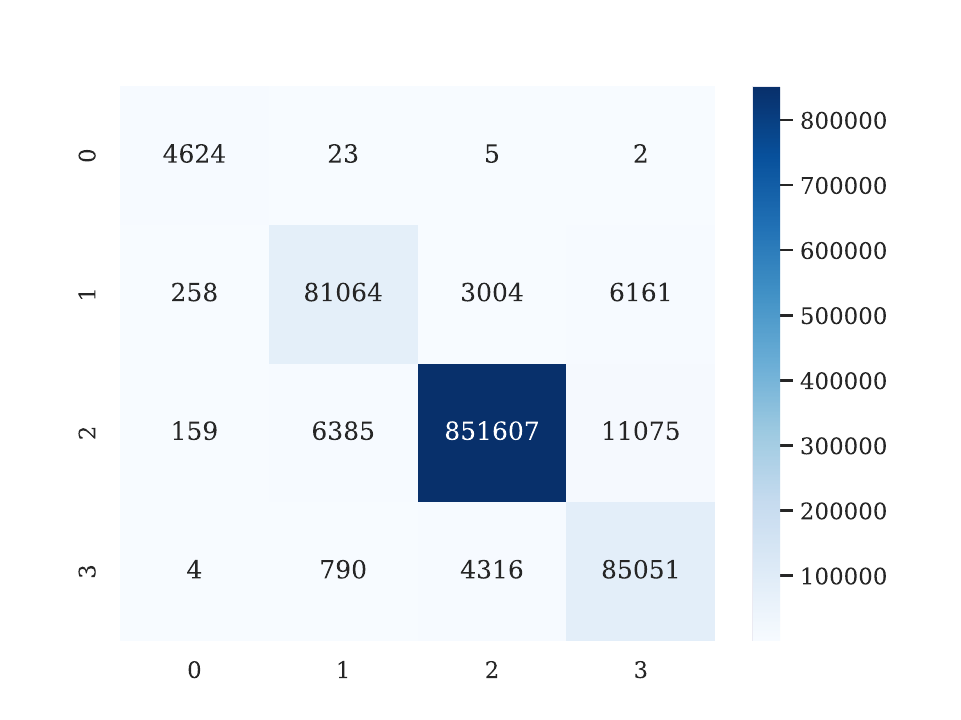}}\label{fig:f1}
		\hfill
		\subfloat[CNN + HMM output.]{\includegraphics[width=0.47\textwidth]{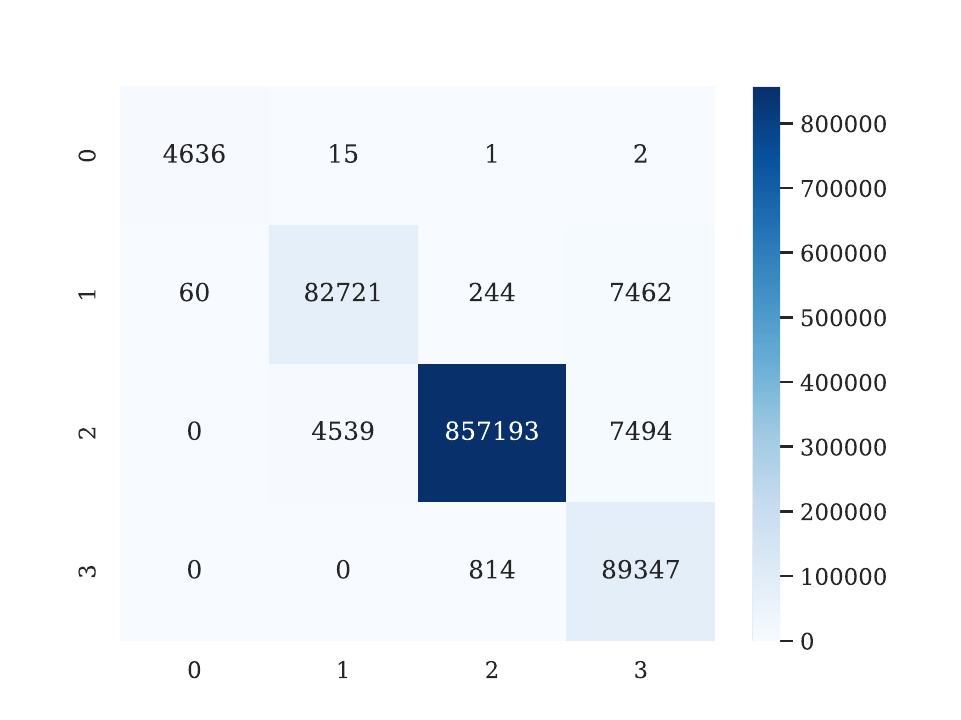}\label{fig:f2}}
		\caption{Confusion matrices of the CNN output (a) and the CNN+HMM combination (b) (classes: esophagus ($0$), stomach ($1$), small intestine~($2$) and colon~($3$)).}
		\label{fig:confusion_hmm}
        \centering
		\subfloat[VCE study ID 94.]{\includegraphics[width=0.47\textwidth]{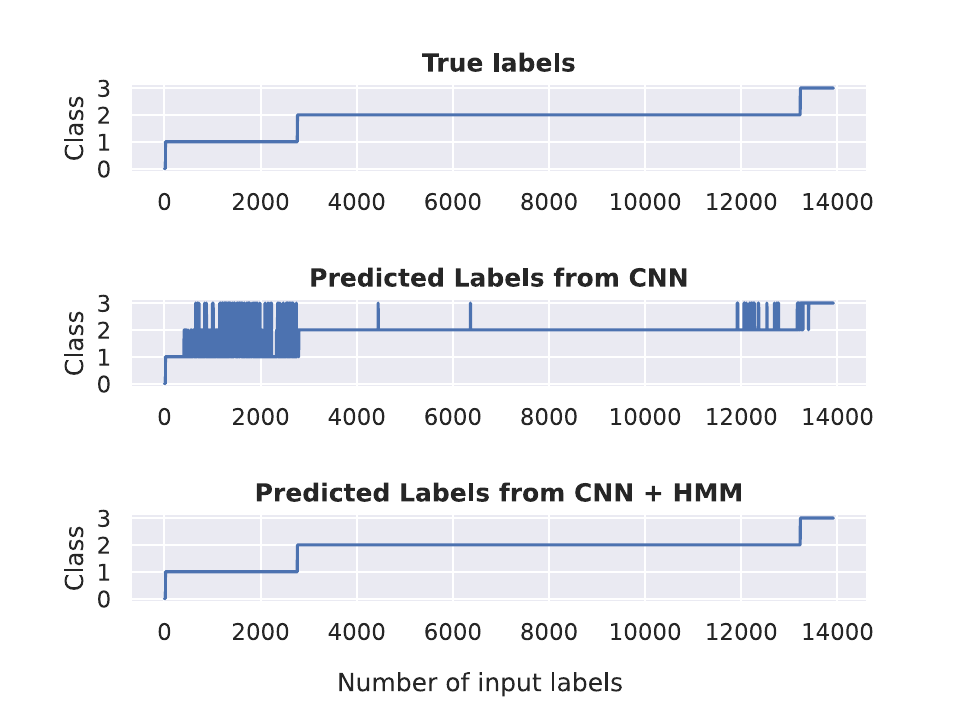}\label{fig:s94}}
		\hfill
		\subfloat[VCE study ID 95.]{\includegraphics[width=0.47\textwidth]{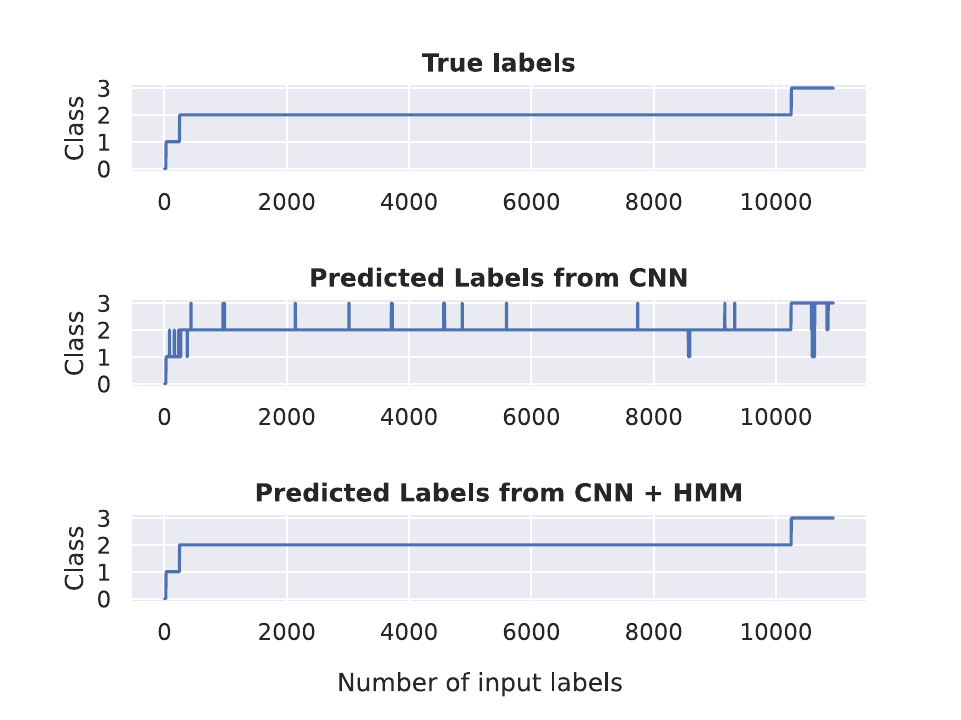}\label{fig:s95}}
		\caption{Comparison of class predictions, exemplarily shown for two VCE studies.}
		\label{fig:comparison_classes}
	\end{figure} 
      
    In all experiments, the MobileNetV3 was trained for $10$ epochs on the RI training set with the AdamW optimizer and a learning rate of $0.001$ with the HANNAH framework~\cite{gerum2022hardware}. The presented approach for localization within the GI tract by combining classification with the CNN and the time-series analysis of the HMM achieved an accuracy of $98.04\%$ with a window size of $w=300$. This is an improvement compared to only applying the MobileNetV3 on its own ($96.95\%$). The corresponding confusion matrices are shown in Figure \ref{fig:confusion_hmm}, displaying the improved classification per class of the combination CNN+HMM (b) and compared to solely using the CNN (a). It becomes apparent that particularly the classification of stomach and colon images was notably enhanced ($3004$ vs. $244$ images misclassified as small intestine and $4316$ vs. $814$ images misclassified as colon). 
    Hence, even with low resolutions high accuracies were achieved (Table~\ref{tab2}) and an overall improved self-localization was demonstrated (Figure~\ref{fig:confusion_hmm}). Importantly, this allows to use low resolutions within the sections outside of the small intestine while still providing a precise classification. Therefore, less energy is required within these sections and the energy can be either used for other tasks or simply leads to a longer battery lifetime. Table~\ref{tab1} extends the results by displaying additional metrics in comparison to the baseline~\cite{charoen2022rhode}. This demonstrates the size reduction of the neural network to $\approx 1$M parameters compared to the baseline model \cite{charoen2022rhode} with $\approx56$M parameters (both computed with 32-bit floating-point values).  

     Exemplarily, in Figure~\ref{fig:comparison_classes}, for two patients, the true labels (top row, corresponding to the perfect solution) of the captured images over time are shown in comparison to the predicted labels from the CNN only (middle row) and finally the predicted labels from the HMM which further processed the output from the CNN (last row). While the CNN still presents some misclassifications, the HMM is able to capture and correct false predictions from the CNN to some extend, resulting in a more similar depiction compared to the true labels.
 
	\begin{table}[h!]
			\caption{Results for the presented approach in comparison to the baseline results (Mean values over all $85$ tested VCE studies).}\label{tab1}\centering
			\begin{tabular}{c|c|c|c}
					\hline
					Metric &  MobilenetV3 + HMM ($w=300$)& MobilenetV3 & Baseline \cite{charoen2022rhode} \\
					\hline
					Accuracy [$\%$] & $98.04$ & $96.95$& $97.1$\\
					Number of Parameters & $\approx1$M & $\approx1$M & $\approx56$M\\
					Averaged MAE &  $0.0350$ & $0.0463$ & -- \\
					Averaged R2-Score & $0.8077$ & $0.7216$ & --\\
					Average Delay (\# Frames) & $19.19$ & --  & --\\
					\hline
				\end{tabular}
		\end{table}

\begin{figure}[h!]
		\includegraphics[width=\textwidth]{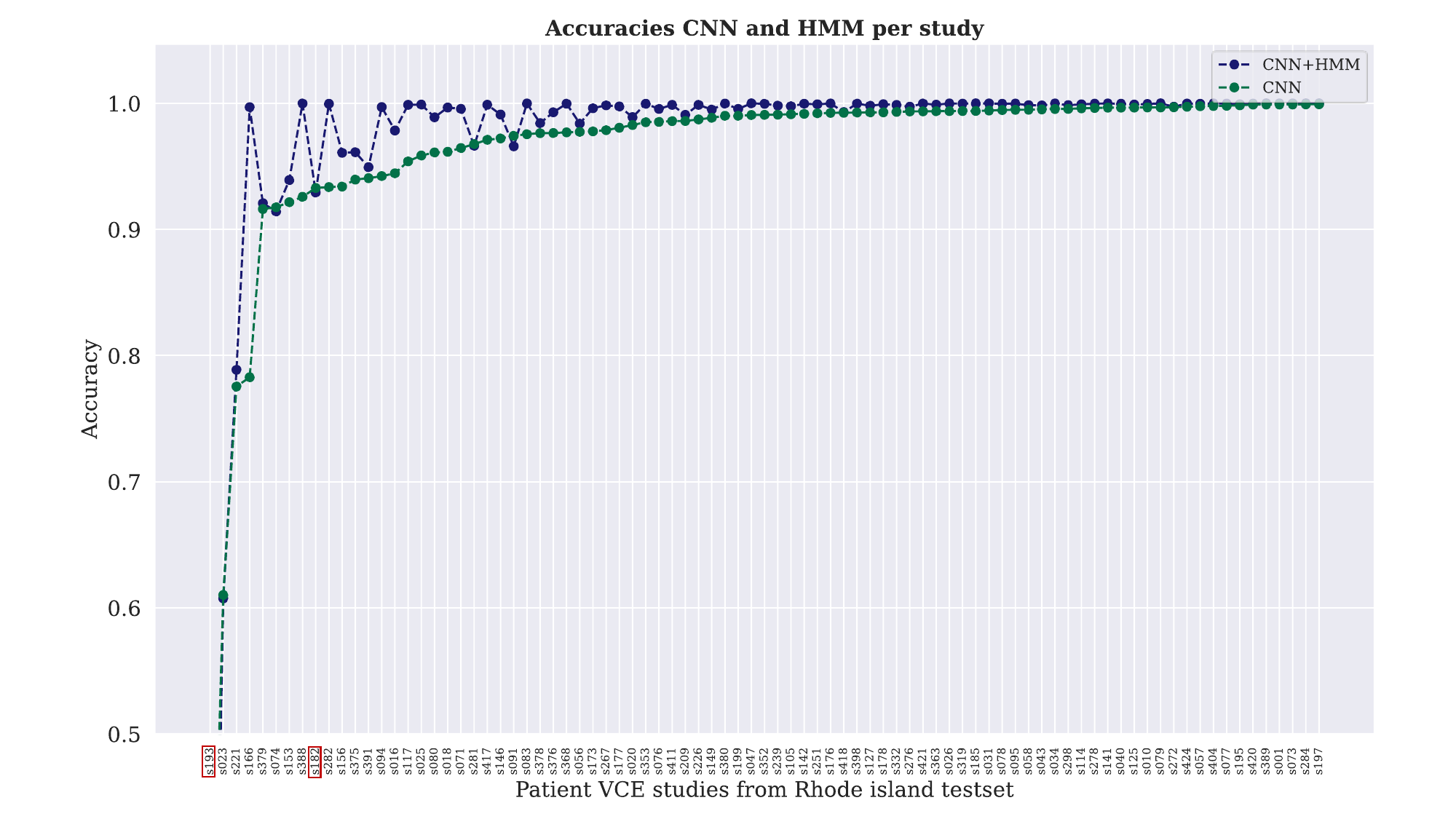}
		\centering
		\caption{Accuracies of the CNN compared to the combinatorial approach CNN+HMM. Marked in red are two studies with worse results if the combination is used, more details can be found in Figure~\ref{fig:s193} and Figure~\ref{fig:s182}.}
		\label{fig:acc_study}
		\centering
		\subfloat[VCE study 193.]{\includegraphics[width=0.47\textwidth]{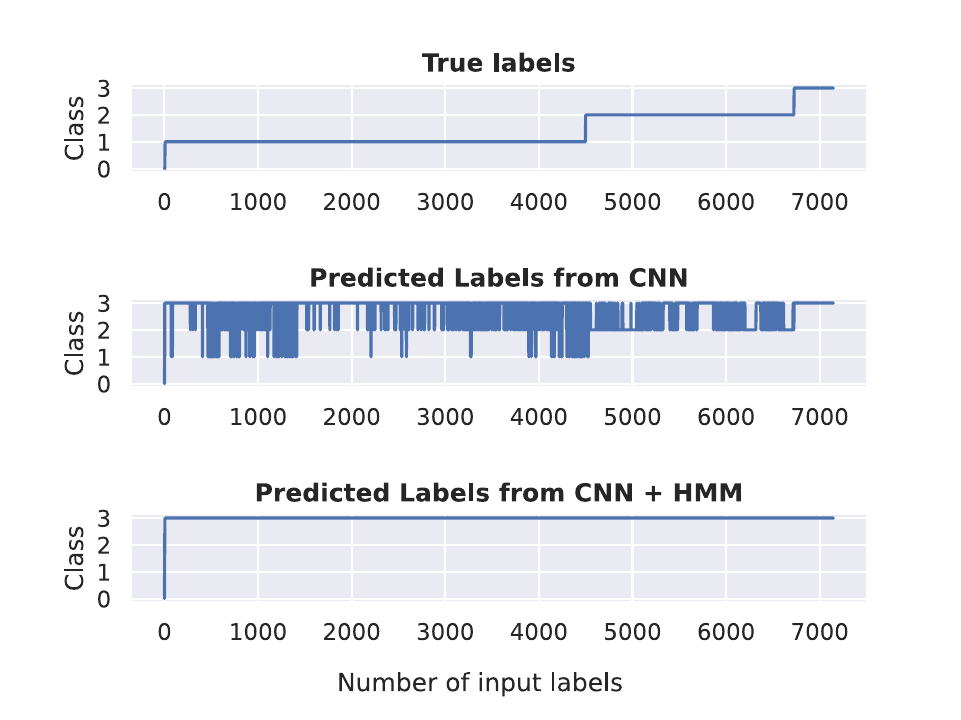}\label{fig:s193}}
		\hfill
		\subfloat[VCE study 182.]{\includegraphics[width=0.47\textwidth]{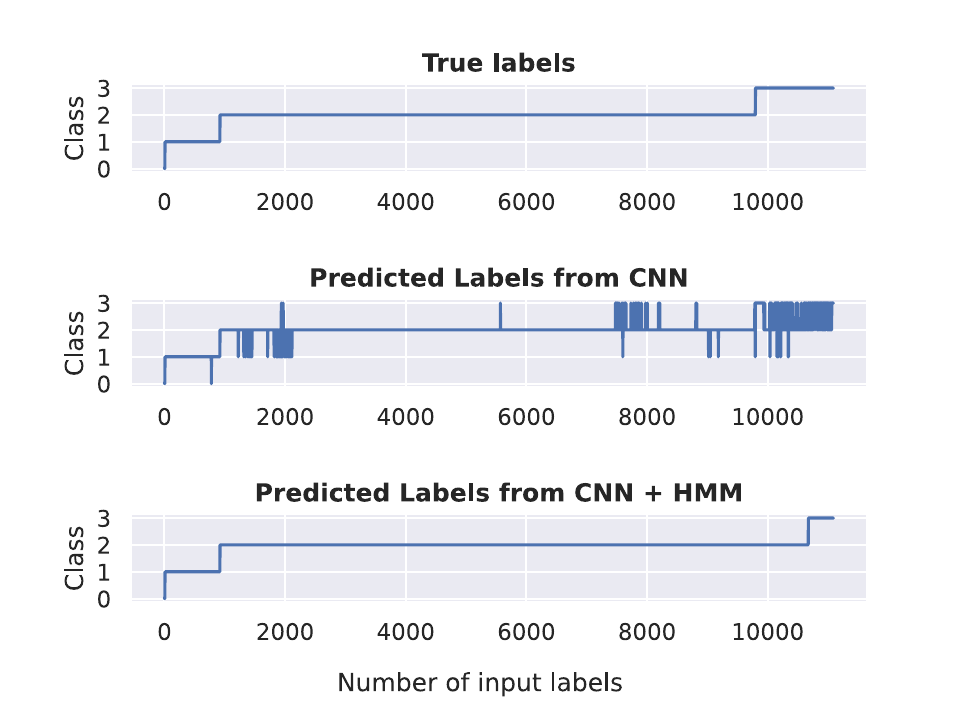}\label{fig:s182}}
		\caption{Comparison of class predictions for two outlier VCE studies.}
		\label{fig:outlier-acc}
	\end{figure}

	The accuracies of class prediction achieved with the CNN compared to the combinatorial approach over all patient VCE studies are visualized in Figure \ref{fig:acc_study}. The CNN+HMM combination achieved superior results compared to the CNN alone for almost all patient studies. Exemplarily, two of the outliers are observed more closely to understand the incidences of worse performance with the combinatorial approach (Figure~\ref{fig:outlier-acc}). Presented in Figure \ref{fig:s193}, one VCE study shows poor results for both approaches. The CNN misclassifies the majority of the images achieving an accuracy of only $25.20\%$. Subsequently, as the preceding classification of the CNN is mostly incorrect, the Viterbi decoding cannot classify the error-prone labels correctly received from the CNN resulting in a very early misclassification as colon. Since the HMM cannot take a step back and the CNN provides such an error-prone output, the remaining images are also classified as colon resulting in an accuracy of $5.93\%$. For one VCE study (Figure~\ref{fig:s182}) overall good results can be achieved, but classification with the combination of CNN+HMM showed a slightly worse result than classifying with the CNN only. It becomes apparent that the CNN has trouble classifying images near the transition of small intestine and colon leading to a delayed transition detection by the HMM. It is expected that a large proportion of misclassifications can be avoided with time-dependent transition probabilities.

	\section{Conclusion}
A pipeline to accurately classify the current gastroenterologic section of VCE images was proposed. The combination of a CNN followed by time-series analysis can automatically classify the present section of the capsule achieving an accuracy of $98.04\%$. Limited by the small size of the embedded device in practical use, considering the given energy constraints is crucial. The presented approach requires only $\approx 1$M parameters while providing a higher accuracy than the current baseline. This approach results in an energy reduction which potentially provides more options on when and how relevant images are captured.

	\newpage
	\bibliographystyle{splncs04}
	\bibliography{references}

\end{document}